%% file: main.tex
\documentclass{article}

\usepackage{spconf}
\input{preamble}

\title{Local Evidence, Global Failure: Diagnosing Speech-Vision Temporal Composition in Audio-Visual Language Models}

\name{
Jaehong Myung$^{1*}$,
Seunghun Oh$^{1*}$,
and Unsang Park$^{1,2\dagger}$
\thanks{$^{*}$Equal contribution.}
\thanks{$^{\dagger}$Corresponding author: Unsang Park (unsangpark@sogang.ac.kr).}
}

\address{
$^{1}$Department of Artificial Intelligence,
Sogang University, Seoul, Republic of Korea\\
$^{2}$Department of Computer Science and Engineering,
Sogang University, Seoul, Republic of Korea
}

\begin{document}
\ninept
\maketitle

\input{sec/00_abstract}
\input{sec/01_intro}
\input{sec/02_method}
\input{sec/03_setup}
\input{sec/04_results}
\input{sec/05_discussion}
\input{sec/06_conclusion}

\clearpage
\section*{Acknowledgments}
This work was supported by the Institute of Information \& Communications
Technology Planning \& Evaluation (IITP) grant funded by the Korea
government (MSIT) (No. 2022-0-00621, RS-2022-II220621,
Development of artificial intelligence technology that provides
dialog-based multi-modal explainability).

\noindent\textbf{Conflict of Interest.}
The authors have no relevant financial or non-financial conflicts of interest to disclose.

\section*{Compliance with Ethical Standards}
This study uses a publicly released benchmark and involves no new collection
of human-participant or animal data.

\bibliographystyle{IEEEbib}
\bibliography{strings,main}

\end{document}

%% file: preamble.tex
\usepackage{amsmath}
\usepackage{amssymb}

\usepackage{graphicx}
\usepackage{booktabs}
\usepackage{multirow}
\usepackage{array}
\usepackage{makecell}
\usepackage{float}
\usepackage{balance}
\usepackage{needspace}

\usepackage{microtype}
\usepackage{hyperref}
\hypersetup{
    hidelinks,
    pdftitle={Local Evidence, Global Failure: Diagnosing Speech-Vision Temporal Composition in Audio-Visual Language Models},
    pdfauthor={Jaehong Myung, Seunghun Oh, Unsang Park},
    pdfsubject={TED manuscript v3.2.6},
    pdfkeywords={speech–vision understanding, audio-visual language models, temporal composition, multimodal hallucination, training-free inference}
}
\usepackage{xcolor}
\usepackage{xspace}
\newcommand{\TED}{\textsc{TED}\xspace}

\newcommand{\TEDCand}{\textsc{TED-Cand}\xspace}
\newcommand{\WHAT}{\textsc{What}\xspace}
\newcommand{\WHEN}{\textsc{When}\xspace}
\newcommand{\pyes}{p_{\mathrm{yes}}}
\newcommand{\smatch}{s_{\mathrm{match}}}
\newcommand{\TF}{\textsc{TF}\xspace}
\newcommand{\TA}{\textsc{TA}\xspace}
\newcommand{\CMTB}{\textsc{CMTB}\xspace}

\usepackage{etoolbox}
\makeatletter
\patchcmd{\@maketitle}{\large \bf}{\fontsize{14}{16}\selectfont\bfseries}{}{\PackageError{TED}{Title-size patch failed}{Check spconf.sty}}
\makeatother

%% file: sec/00_abstract.tex
\begin{abstract}
Audio-visual large language models (AV-LLMs) can fail at full-video speech--vision decisions even when useful visual evidence is locally recoverable. We introduce Temporal Evidence Decomposition (\TED), a training-free diagnostic framework that separates local visual verification from speech-relative aggregation. Automatic speech recognition supplies a transcript and utterance interval. \TED queries audio-free windows with the transcript for Temporal Alignment and Forecasting, or answer candidates for Cross-Modal Temporal Binding, then aggregates evidence relative to that interval. Across four frozen AV-LLMs on SVHalluc, Temporal Forecasting accuracy rises from 29.3--33.7\% under full-video questioning to 79.3--82.2\% with \TED. Query replacement reduces recovery; the observed anchor remains beneficial when temporal-region availability is held fixed. Physical-isolation and timing-shortcut controls test alternative explanations. The findings distinguish locally recoverable evidence from successful full-video decisions, motivating explicit separation of narration content, utterance timing, and time-indexed visual evidence.
\end{abstract}

\begin{keywords}
speech--vision understanding, audio-visual language models, temporal composition, multimodal hallucination, training-free inference
\end{keywords}

%% file: sec/01_intro.tex
\section{Introduction}
\label{sec:intro}

Narration need not describe what is currently visible. In a cooking video, for example, a speaker may describe seasoning while ingredients are still being chopped, or refer to a completed step. Understanding this mismatch requires relating the narrated action's visual occurrence to the utterance time. Speech supplies distinct cues: \emph{what} is said and \emph{when}.

SVHalluc evaluates this distinction through three temporal tasks~\cite{zhang2026svhalluc}. Temporal Alignment (\TA) asks whether the action described in speech is visible during the utterance. Temporal Forecasting (\TF) asks whether the narrated action occurs before, during, or after the utterance. Cross-Modal Temporal Binding (\CMTB) asks which candidate action is visible during the utterance, even when the narration describes a different action. Open-source audio-visual large language models (AV-LLMs) perform near chance on \TA/\TF, despite strong speech recognition and utterance localization in SVHalluc's Qwen3 analysis. Complementary speaker-centric evaluation also tests who speaks, what is said, and when it occurs~\cite{nguyen2026see}. A wrong final answer, however, does not distinguish missing visual evidence from difficulty using that evidence relative to speech.

\begin{figure*}[t]
    \centering
    \includegraphics[width=\textwidth]{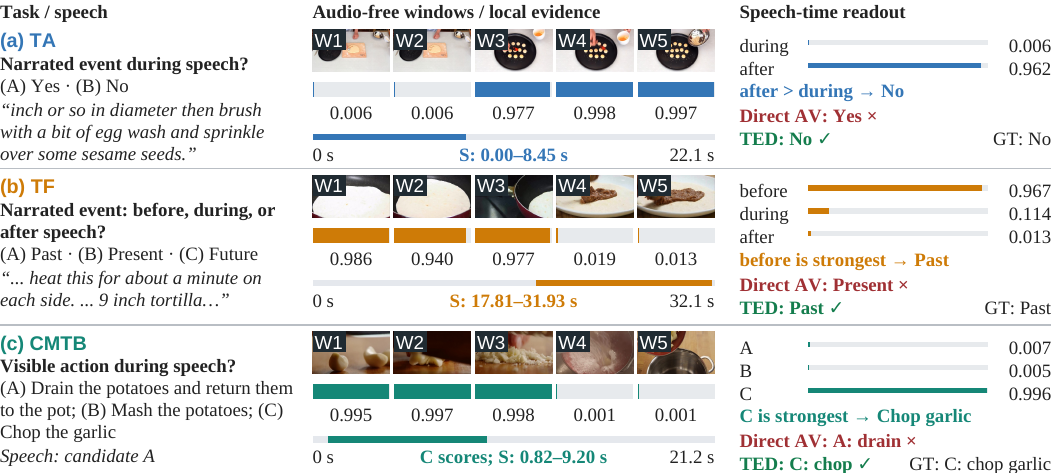}
    \caption{Illustrative Qwen3 traces from SVHalluc. Center-frame thumbnails represent full audio-free windows. Task questions and transcript excerpts are shortened for display only; scoring uses the original queries. \TA/\TF use transcript queries; \CMTB scores each answer candidate (window scores shown for C). Right-hand bars aggregate evidence relative to the ASR-derived utterance interval $S$. Values are individual traces, not averages.}
    \label{fig:examples}
\end{figure*}

\emph{Temporal Evidence Decomposition} (\TED) compares two interfaces to the same frozen backbone. \emph{Direct} receives the full video, speech-bearing audio, and benchmark question in one call. In \TED, a text query (\WHAT) specifies what to verify: the transcript from automatic speech recognition (ASR) for \TA/\TF, or an answer candidate for \CMTB. The frozen backbone scores the query in audio-free windows; an external readout aggregates these scores relative to the ASR-derived utterance interval, the temporal anchor (\WHEN). Work on grounded video question answering (VideoQA) shows that correct answers need not rest on relevant visual evidence~\cite{xiao2024can}. We test the complementary possibility: useful local evidence despite an incorrect full-video answer (Fig.~\ref{fig:examples}). This deliberately changes the inference interface while retaining the categorical task.

Speech-aware AV-LLMs~\cite{sun2024video} and fine-grained audiovisual grounding methods~\cite{chowdhury2024meerkat,kim2025qatiger} seek richer multimodal representations. Hallucination diagnostics~\cite{guan2024hallusionbench,sung2025avhbench,seth2026exploring} and decoding-based mitigation~\cite{jung2026avcd} examine the reliability of multimodal answers. Temporal video-language models~\cite{ren2024timechat,huang2024lita,wang2025groundedvideollm} and long-context benchmarks~\cite{wu2024longvideobench} address temporal representation and evidence retrieval. Closest to our verification step, local-confidence methods use external aggregation for temporal grounding~\cite{aklilu2025zeal,li2026reze,huang2026vlmknows}, and attention analyses expose evidence not reliably used in timestamp generation~\cite{du2026mllmsknow}. We use these local-verification primitives to diagnose speech-relative decisions, not to introduce a new scanning rule.

Across the four evaluated AV-LLMs (Table~\ref{tab:main}), poor full-video accuracy coexists with useful local evidence. Recovery is query-specific and benefits from the observed utterance time even with unchanged candidate-region availability; physical-isolation and timing controls examine alternative explanations.

\begin{table*}[t]
\centering
\caption{Direct AV inference versus Matched \TED on the three SVHalluc temporal tasks (\%). \TA also reports positive-class precision, recall, and F1. Bold highlights higher accuracy within each backbone.}
\label{tab:main}
\fontsize{9}{10.2}\selectfont
\setlength{\tabcolsep}{2.2pt}
\renewcommand{\arraystretch}{1.04}
\begin{tabular*}{\textwidth}{@{\extracolsep{\fill}}llcccccc@{}}
\toprule
\multirow{2}{*}{Backbone} & \multirow{2}{*}{Mode}
& \multicolumn{4}{c}{TA} & TF & CMTB \\
\cmidrule(lr){3-6}\cmidrule(lr){7-7}\cmidrule(lr){8-8}
& & Acc. $\uparrow$ & Prec. $\uparrow$ & Recall $\uparrow$ & F1 $\uparrow$ & Acc. $\uparrow$ & Acc. $\uparrow$ \\
\midrule
\multirow{2}{*}{Qwen3-Omni-30B~\cite{xu2025qwen3omni}}
& Direct & 47.55 & 49.13 & 91.85 & 64.02 & 29.35 & 55.46 \\
& \TED & \textbf{77.99} & 92.62 & 61.41 & 73.86 & \textbf{80.43} & \textbf{84.66} \\
\cmidrule(lr){1-8}
\multirow{2}{*}{Qwen2.5-Omni-7B~\cite{xu2025qwen25omni}}
& Direct & 53.80 & 64.58 & 16.85 & 26.72 & 33.70 & 51.92 \\
& \TED & \textbf{77.45} & 93.16 & 59.24 & 72.43 & \textbf{79.35} & \textbf{86.14} \\
\cmidrule(lr){1-8}
\multirow{2}{*}{VideoLLaMA2.1-7B-AV~\cite{cheng2024videollama2}}
& Direct & 50.00 & 50.00 & 100.00 & 66.67 & 30.80 & 49.85 \\
& \TED & \textbf{75.54} & 91.23 & 56.52 & 69.80 & \textbf{79.35} & \textbf{83.78} \\
\cmidrule(lr){1-8}
\multirow{2}{*}{video-SALMONN~2+ 7B~\cite{tang2025videosalmonn2}}
& Direct & 49.73 & 49.86 & 99.46 & 66.42 & 33.33 & 48.97 \\
& \TED & \textbf{79.35} & 94.31 & 63.04 & 75.57 & \textbf{82.25} & \textbf{88.20} \\
\bottomrule
\end{tabular*}
\end{table*}

%% file: sec/02_method.tex
\section{Temporal Evidence Decomposition}
\label{sec:method}

\subsection{Query-conditioned local evidence}
Whisper~\cite{radford2023whisper} supplies transcript $q_s$ and timestamped speech segments. Their convex hull defines the observed utterance interval $S=[s,e)$, from the first segment start to the last segment end. This is a speech-time anchor, not a visual-event annotation. We divide the video of duration $T$ into $N=5$ uniform, audio-free windows $W_i=[a_i,b_i)$. A frozen verifier assigns each window a query-conditioned score $z_i(q)$.

A generative verifier is asked whether the window visually shows the queried content, using both $(A)$ Yes/$(B)$ No and the reversed order. Let $\ell_{i,\rm yes}^{(o)}$ and $\ell_{i,\rm no}^{(o)}$ be first-token log-probabilities of the letters corresponding to Yes and No under order $o$. We compute
\begin{equation}
 p_i^{(o)}=\frac{e^{\ell_{i,\rm yes}^{(o)}}}{e^{\ell_{i,\rm yes}^{(o)}}+e^{\ell_{i,\rm no}^{(o)}}},\quad
 \pyes(W_i,q)=\tfrac{1}{2}\sum_o p_i^{(o)}.
 \label{eq:pyes}
\end{equation}
We use $z_i(q)=\pyes(W_i,q)$ as comparative evidence, not a calibrated event probability.

For SigLIP~2~\cite{tschannen2025siglip2}, $z_i(q)=\smatch(W_i,q)$ averages frame--text compatibility. Cosine similarities of normalized embeddings pass through the learned logit scale and bias and a sigmoid before averaging. 

\subsection{Narration-conditioned \TED}
For \TA/\TF, the transcript supplies \WHAT and $S$ supplies \WHEN. A transcript may describe multiple actions, so $z_i(q_s)$ measures compatibility with that span. The reference interval $S$ defines regions before, during, and after it. For $r\in\{\mathrm{before},\mathrm{during},\mathrm{after}\}$, let $\omega_{ir}$ be the overlap duration of window $W_i$ and region $r$. We compute
\begin{equation}
 E_r(q)=\frac{\sum_i \omega_{ir}z_i(q)}{\sum_i\omega_{ir}}.
 \label{eq:region}
\end{equation}
A zero-overlap region is unavailable. \TF returns the benchmark label of the region with the largest available $E_r(q_s)$. \TA requires the during region and at least one alternative: it predicts Yes if $E_{\mathrm{during}}(q_s)$ exceeds all available alternatives, and No if any exceeds it. Ties or missing required evidence yield abstention.

\emph{Matched} \TED uses the target transcript and observed anchor. For the \TF controls, Other and Generic replace only the query with a donor transcript or a non-specific cooking query, respectively. The anchor controls keep the target query and window scores fixed and replace the reference interval without changing the benchmark labels.

Anchor replacement can remove candidate regions: an interval starting at zero leaves no region before it. We therefore compare two \TF anchor controls. Unrestricted shuffled-\WHEN (Unrestr.-Shuf.) normalizes a donor's interval endpoints by its video duration and scales them to the target duration. Support-preserving shuffled-\WHEN (SP-Shuf.) starts from this map, retaining valid assignments that leave each target's set of non-empty regions unchanged. It deterministically reassigns remaining donors one-to-one without self-donors, minimizing change from the original shuffled intervals under that constraint. SP-Shuf. is a constrained repair, not a fresh random permutation.

\subsection{Candidate-conditioned \TED}
For \CMTB, each answer candidate $q_k$ supplies \WHAT; speech supplies only \WHEN. \TEDCand selects the candidate with the largest $E_{\mathrm{during}}(q_k)$. To test the role of the utterance interval, Shuffled retains all candidate--window scores and transfers an unrestricted donor anchor. The no-\WHEN control averages each candidate's scores over all five windows.

 Both narration- and candidate-conditioned \TED use frozen verifiers and fixed readouts, without ground-truth visual-event intervals or threshold tuning.

%% file: sec/03_setup.tex
\section{Experimental Setup}
\label{sec:setup}

\noindent\textbf{Benchmark and evaluation.}
We use the public SVHalluc temporal splits: 368 \TA, 276 \TF, and 339 \CMTB instances. Their option-balanced labels give chance accuracies of 50.0\%, 33.3\%, and 33.3\%, respectively. We report accuracy for all tasks and Yes-class precision, recall, and F1 for \TA. Missing, failed, incomplete, and abstained outputs remain errors on the full evaluation sets.

\par\noindent\textbf{Verifiers.}
We evaluate Qwen3-Omni-30B-A3B (Instruct)~\cite{xu2025qwen3omni}, Qwen2.5-Omni-7B~\cite{xu2025qwen25omni}, VideoLLaMA2.1-7B-AV~\cite{cheng2024videollama2}, and video-SALMONN~2+~7B~\cite{tang2025videosalmonn2}. SigLIP~2 So400m-\allowbreak384~\cite{tschannen2025siglip2} supplies contrastive scores to the same \TF/\CMTB readouts. SigLIP~2 uses its native 64-token text limit, truncating longer transcripts.

\par\noindent\begin{minipage}{\linewidth}
Direct scores are re-evaluated in our implementation; Qwen3 video-decoding failures remain errors. For Qwen2.5 Direct, only pre-generation video-decoding failures trigger a single retry with an alternate decoder; all evaluation items are retained.
\end{minipage}

\begin{figure}[H]
    \centering
    \includegraphics[width=\columnwidth]{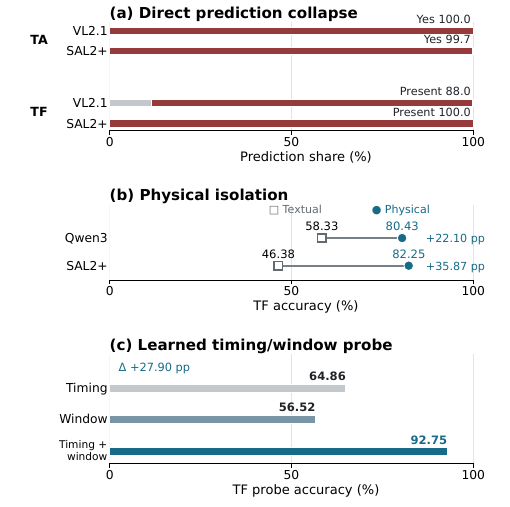}
    \caption{Diagnostics of full-video failure and local recovery. \textbf{(a)} Direct prediction distributions. \textbf{(b)} Single-run, matched-call comparison of local clips and full-video interval prompts; temporal coverage is matched for SAL2+ and approximate for Qwen3. \textbf{(c)} Learned timing/window probe under source-video-grouped cross-validation. Panel (c) shows descriptive out-of-fold probe accuracies, not \TED scores.}
    \label{fig:diagnostics}
\end{figure}

\par\noindent\textbf{Decoding and controls.}
Direct uses greedy answer decoding. Generative \TED uses 10 local calls per \TA/\TF item and per \CMTB candidate, in addition to ASR. A donor mapping assigns another item's transcript or interval to each target. Query and anchor mappings are separate; each control uses five mappings shared across verifiers. Mappings exclude self-donors and use neither ground-truth labels nor verifier scores. Donor-dependent metrics report mean and sample standard deviation (SD). Paired accuracy comparisons use exact two-sided McNemar tests, separately per mapping for donor controls, without pooling items across mappings or averaging $p$-values.

\Needspace{14\baselineskip}

\par\noindent\textbf{Additional diagnostics.}
Direct prediction distributions use full splits; answer-option permutations and question paraphrasing use balanced \TA ($n=100$) and \TF ($n=99$) subsets for VideoLLaMA2.1 and video-SALMONN~2+. On Qwen3 \TF, ASR-assisted full video answers the benchmark question in one call from an audio-free full video, the full ASR transcript, and segment timestamps, testing explicit access to the information forming \TED's query and anchor. Physical isolation compares local clips with repeated full-video interval prompts at matched call counts on Qwen3 and video-SALMONN~2+ \TF. Non-boundary and segment-union analyses reuse stored outputs. Audio-reintroduction and frame-order controls characterize the recovered evidence.

\par\noindent\textbf{Learned timing/window probe.}
On the full 276-item \TF split, we test whether local scores add predictive information beyond specified timing features. Timing features describe the ASR interval and video duration; window features are Qwen3's five transcript-conditioned scores and summary statistics. Separate multinomial logistic regressions (fixed $C=1$, no hyperparameter search) predict \TF labels from either feature set or both. Five-fold cross-validation groups items by source video; standardization uses training folds only. The descriptive out-of-fold (OOF) accuracies concern this learned probe, not \TED.

%% file: sec/04_results.tex
\section{Results and Analysis}
\label{sec:results}

We examine local recovery, query and anchor contributions, and alternative explanations.

\subsection{Recoverable local evidence}
Table~\ref{tab:main} shows Direct-to-\TED accuracy increases across four AV-LLMs: 47.6--53.8\% to 75.5--79.3\% on \TA, 29.3--33.7\% to 79.3--82.2\% on \TF, and 49.0--55.5\% to 83.8--88.2\% on \CMTB. The Qwen3 \TF gain is 51.09 percentage points (pp; exact McNemar $p=2.62\!\times\!10^{-32}$). Thus, the same frozen backbones supply useful evidence under local queries despite unreliable full-video answers.

\Needspace{8\baselineskip}
Near-chance \TA/\TF scores conceal response preferences: 87.0--100\% of \TA predictions and 62.7--100\% of \TF predictions fall on one semantic label per backbone, without comparable \CMTB collapse (examples in Fig.~\ref{fig:diagnostics}(a)). In fixed-subset Direct tests for VideoLLaMA2.1 and video-SALMONN~2+, answer-option permutations and question paraphrasing preserve the dominant semantic labels, with no significant accuracy change (all exact $p\geq0.45$). The preference follows label meaning, not answer-letter position.



\begin{table}[H]
\centering
\caption{Query and anchor controls (accuracy, \%). Donor-dependent entries are mean $\pm$ sample SD over five mappings; others show one value. \textbf{(a)} Matched \TED uses the target transcript and observed anchor; replacements change only WHAT or WHEN. SP-Shuf. preserves the available-region set. \textbf{(b)} \CMTB anchor controls; C$-$S is Correct minus Shuffled (pp).}
\label{tab:controls}
\fontsize{9}{10.2}\selectfont
\renewcommand{\arraystretch}{1.02}

{\centering\textbf{(a) Narration-conditioned \TF}\par}
\vspace{1pt}
\begingroup
\fontsize{9}{10.5}\selectfont
\renewcommand{\arraystretch}{1.0}
\setlength{\tabcolsep}{1.0pt}

\newcommand{\TEDtableSD}[1]{\textpm#1}

\begin{tabular*}{\columnwidth}{@{\extracolsep{\fill}}lccccc@{}}
\toprule
\multirow{2}{*}{Verifier}
& \multicolumn{1}{c}{\TED}
& \multicolumn{2}{c}{WHAT replacement}
& \multicolumn{2}{c}{WHEN replacement} \\
\cmidrule(lr){2-2}\cmidrule(lr){3-4}\cmidrule(lr){5-6}
& Matched & Other & Generic & SP-Shuf. & \makecell{Unrestr.-\\Shuf.} \\
\midrule
Qwen3
& \textbf{80.43}
& 42.46\TEDtableSD{2.97}
& 38.04
& 71.81\TEDtableSD{1.24}
& 54.13\TEDtableSD{2.40} \\

Qwen2.5
& \textbf{79.35}
& 41.09\TEDtableSD{3.03}
& 35.14
& 71.74\TEDtableSD{2.13}
& 53.12\TEDtableSD{2.36} \\

VL2.1
& \textbf{79.35}
& 41.59\TEDtableSD{2.94}
& 36.23
& 68.41\TEDtableSD{1.10}
& 50.72\TEDtableSD{2.84} \\

SAL2+
& \textbf{82.25}
& 42.54\TEDtableSD{2.96}
& 36.59
& 70.43\TEDtableSD{1.51}
& 51.52\TEDtableSD{3.21} \\

SigLIP2
& \textbf{73.91}
& 41.96\TEDtableSD{1.75}
& 40.22
& 66.59\TEDtableSD{1.63}
& 50.36\TEDtableSD{2.86} \\
\bottomrule
\end{tabular*}\par
\endgroup

\vspace{3pt}
{\centering\textbf{(b) Candidate-conditioned \CMTB}\par}
\vspace{1pt}

\begingroup
\fontsize{9}{10.5}\selectfont
\renewcommand{\arraystretch}{1.0}
\setlength{\tabcolsep}{1.0pt}

\newcommand{\TEDtableSD}[1]{\textpm#1}

\begin{tabular*}{\columnwidth}{@{\extracolsep{\fill}}lcccc@{}}
\toprule
Verifier & Correct & Shuffled & no-WHEN & C$-$S \\
\midrule
Qwen3~\cite{xu2025qwen3omni}
& \textbf{84.66}
& 48.26\TEDtableSD{2.07}
& 51.92
& +36.40\TEDtableSD{2.07} \\

Qwen2.5~\cite{xu2025qwen25omni}
& \textbf{86.14}
& 46.61\TEDtableSD{1.83}
& 43.95
& +39.53\TEDtableSD{1.83} \\

VL2.1~\cite{cheng2024videollama2}
& \textbf{83.78}
& 47.26\TEDtableSD{2.04}
& 45.13
& +36.52\TEDtableSD{2.04} \\

SAL2+~\cite{tang2025videosalmonn2}
& \textbf{88.20}
& 46.84\TEDtableSD{2.46}
& 49.85
& +41.36\TEDtableSD{2.46} \\

SigLIP2~\cite{tschannen2025siglip2}
& \textbf{74.63}
& 45.31\TEDtableSD{1.79}
& 44.54
& +29.32\TEDtableSD{1.79} \\
\bottomrule
\end{tabular*}\par
\endgroup
\end{table}

\subsection{Query and anchor contributions}
Table~\ref{tab:controls}(a) tests query and anchor contributions. Across all five verifiers, Other and Generic reduce accuracy by 32.0--45.7 pp relative to Matched \TED. This query sensitivity also holds for the non-generative verifier.

SP-Shuf. lowers accuracy by 7.3--11.8 pp while preserving each item's available regions. The observed anchor remains beneficial when the candidate-region set is held fixed. Unrestr.-Shuf. yields a 23.6--30.7 pp drop, but also makes the ground-truth region unavailable for 27--32 of 276 items per mapping, forcing errors on those items. Because the two shuffles allow different donor anchors, their accuracy difference does not isolate the effect of removing candidate regions.

For \CMTB, Correct exceeds Shuffled by 29.3--41.4 pp; all 25 verifier--mapping comparisons are significant (exact $p\leq1.50\times10^{-23}$). Separate Shuffled--no-\WHEN comparisons are not significant (all exact $p\geq0.052$), which does not establish equivalence. The Correct--Shuffled advantage supports using the observed utterance interval for candidate selection.

\subsection{Alternative explanations and evidence characteristics}
The ASR-assisted full-video control reaches 32.97\% on Qwen3 \TF, versus 80.43\% for \TED. Explicit transcript and timestamp access therefore does not reproduce the recovery obtained with local verification and external aggregation in this setting.

The physical-isolation results in Fig.~\ref{fig:diagnostics}(b) show 82.25\% with local clips versus 46.38\% with repeated full-video interval prompts for video-SALMONN~2+, under matched call counts and coverage. Qwen3 shows 80.43\% versus 58.33\% with approximately matched coverage. Physically restricting input thus contributes beyond repeated local judgments on the full video.

Under the observed anchor, 89/276 \TF items have no region before the utterance, and none is labeled Past. On the remaining 187 non-boundary items, \TED accuracy is 72.19--77.01\% across the four AV-LLMs, versus 79.35--82.25\% on the full split. All four retain Direct-to-\TED gains of 37.4--53.5 pp (exact $p\leq4.45\times10^{-9}$). This subset is not class-balanced: balanced accuracy is 29.6--33.9\% for Direct and 73.6--77.2\% for \TED. Recovery is not confined to boundary items.

On the full 276-item \TF split, the learned timing/window probe in Fig.~\ref{fig:diagnostics}(c) achieves 64.86\% with timing features, 56.52\% with window features, and 92.75\% with both. The 27.90 pp improvement indicates additional predictive value beyond the specified timing features in this fixed, source-video-grouped evaluation. These are descriptive OOF accuracies of the learned probe, not \TED accuracies.

Reintroducing speech-bearing audio in single-run \TF controls lowers accuracy by 8--20 pp for Qwen3, VideoLLaMA2.1, and video-SALMONN~2+. For Qwen3, frame shuffling and reversal yield 79.71\% and 78.99\%, compared with 80.43\% in the original order. This favors a local visual-state interpretation over a claim of fine-grained motion-order understanding.

%% file: sec/05_discussion.tex
\section{Discussion}
\label{sec:discussion}

\noindent\textbf{Interpretation.}
Query-specific recovery and an anchor benefit with unchanged region availability identify conditions under which local evidence supports correct decisions. The intervention changes the input representation, call count, and readout together, so it cannot identify a unique internal mechanism. Useful evidence thus remains accessible through local verification even when the original full-video interface fails.

\par\noindent\textbf{Scope and limitations.}
\TED uses ASR-derived timing and, for \TA/\TF, transcript-based queries rather than directly grounding acoustic speech. Transcripts can contain multiple actions, and SVHalluc provides no visual-event intervals for testing precise localization. Boundary structure remains informative under SP-Shuf.; preserving region availability does not remove every timing shortcut. Replacing the utterance hull with the union of ASR segments changes accuracy by at most 1.09 pp across 14 task--verifier conditions, a sensitivity result rather than an equivalence test. These findings concern the evaluated checkpoints on SVHalluc.

\par\noindent\textbf{Design implication.}
The results motivate retaining time-indexed evidence and exposing utterance time separately to the readout, distinguishing narrated from concurrently visible actions. Whether this separation improves an end-to-end model remains to be tested.

%% file: sec/06_conclusion.tex
\section{Conclusion}
\label{sec:conclusion}
\TED shows that unreliable full-video speech--vision answers can coexist with useful local evidence from the same frozen backbones. Query replacement reduces recovery, and the observed utterance time remains beneficial even at fixed region availability. Physical-isolation and non-boundary analyses further delimit this result. The contribution is a diagnosis of the gap between recoverable evidence and its use in speech-relative decisions, motivating systems that preserve time-indexed visual evidence and explicitly relate it to narration content and timing.